\documentclass{article} 
\usepackage{iclr2026_conference,times}

\usepackage{amsmath,amsfonts,bm}

\def\eqref#1{equation~\ref{#1}}

\def\1{\bm{1}}

\DeclareMathAlphabet{\mathsfit}{\encodingdefault}{\sfdefault}{m}{sl}
\SetMathAlphabet{\mathsfit}{bold}{\encodingdefault}{\sfdefault}{bx}{n}

\usepackage{hyperref}
\usepackage{url}
\usepackage{graphicx}
\usepackage{caption}
\usepackage{subcaption}

\usepackage{multirow}

\usepackage{wrapfig}
\usepackage[table]{xcolor}
\usepackage{colortbl}
\usepackage{nicematrix}
\usepackage{subcaption}
\usepackage{caption}

\usepackage{pifont}
\usepackage{booktabs}

\usepackage[nottoc]{tocbibind}
\usepackage{minitoc}
\usepackage{bbm}

\title{Learning Skills from Action-Free Videos}

\iclrfinalcopy

\author{Hung-Chieh Fang$^{1}$\footnote[1]{Equal contribution} , Kuo-Han Hung$^{1}$\footnote[1]{Equal contribution} , 
Chu-Rong Chen$^{1}$, Po-Jung Chou$^{1}$ \And  Chun-Kai Yang$^{1}$,  Po-Chen Ko$^{1}$, Yu-Chiang Frank Wang$^{1,2}$, Yueh-Hua Wu$^{2}$ \And  Min-Hung Chen$^{2}$, Shao-Hua Sun$^{1}$ \\
\AND 
$^{1}$ \textnormal{National Taiwan University} \And 
$^{2}$ \textnormal{NVIDIA}
}

\newcommand{\sun}[1]{{\color{blue}{\small\bf\sf [Sun: #1]}}}

\usepackage[normalem]{ulem}

\newcommand{\Skip}[1]{}

\newcommand{\dotieconcat}[2]{
  \text{\raisebox{.8ex}{$\smallfrown$}}%
}

\makeatletter
\newcommand\dslfontsize{\@setfontsize\dslfontsize\@viipt\@viiipt}
\makeatother

\newcommand{\myparagraph}[1]{\noindent \textbf{#1.}}

\usepackage{color}
\usepackage{xcolor}
\usepackage{epsfig,epsf,graphicx,graphics}
\usepackage{caption}
\usepackage{subcaption}
\usepackage[]{mdframed}
\usepackage{tcolorbox}
\usepackage{caption} 
\usepackage{xspace}

\definecolor{codegreen}{rgb}{0,0.6,0}
\definecolor{codegray}{rgb}{0.3,0.3,0.3}
\definecolor{codepurple}{rgb}{0.58,0,0.82}
\definecolor{backcolour}{rgb}{0.95,0.95,0.92}

\usepackage{algorithm}
\usepackage{algorithmic}
\usepackage{xcolor}

\usepackage{courier}

\definecolor{darkred}{rgb}{0.6,0.0,0.0}
\definecolor{darkgreen}{rgb}{0,0.50,0}
\definecolor{lightblue}{rgb}{0.0,0.42,0.91}
\definecolor{orange}{rgb}{0.99,0.48,0.13}
\definecolor{grass}{rgb}{0.18,0.80,0.18}
\definecolor{pink}{rgb}{0.97,0.15,0.45}

\usepackage{minted}
\usepackage[smartEllipses]{markdown}

\def\Method{\texttt{SOF}}
\def\FullMethod{{Skill Abstraction from Optical Flow}}
\newcommand{\numerr}[2]{#1 \scriptsize \textcolor{gray}{$\pm$ #2}}

\newcommand{\panda}[1]{#1}
\newcommand{\sawyer}[1]{#1}

\begin{document}
\doparttoc 
\faketableofcontents 

\maketitle

\begin{abstract}

Learning from videos offers a promising path toward generalist robots by providing rich visual and temporal priors beyond what real robot datasets contain. While existing video generative models produce impressive visual predictions, they are difficult to translate into low-level actions. Conversely, latent-action models better align videos with actions, but they typically operate at the single-step level and lack high-level planning capabilities. We bridge this gap by introducing \FullMethod{} (\Method{}), a framework that learns latent skills from large collections of action-free videos. Our key idea is to learn a latent skill space through an intermediate representation based on optical flow that captures motion information aligned with both video dynamics and robot actions. By learning skills in this flow-based latent space, \Method{} enables high-level planning over video-derived skills and allows for easier translation of these skills into actions. Experiments show that our approach consistently improves performance in both multitask and long-horizon settings, demonstrating the ability to acquire and compose skills directly from raw visual data.

\end{abstract}
\section{Introduction}
\label{sec:intro}
Learning from videos has become a promising direction for scaling up data collection toward generalist robots~\citep{mccarthy2024survey}. These large and diverse video datasets naturally capture the physical dynamics of the world and demonstrate how to complete tasks across a wide range of environments -- capabilities that go beyond traditional robot data, which is often difficult to collect and lacks diversity. Prior work has explored leveraging such videos by learning video models for planning~\citep{du2023learning, ko2023learning, mendonca2023structured, 
 ajay2023compositional,
yang2024learning, videoworldsimulators2024, zhou2024robodreamer, 
liang2024dreamitate,luo2025grounding}, reward models for reinforcement learning that infer task progress~\citep{nair2022r3m, ma2023liv, yang2024rank2reward, hung2025victor}, representations for downstream policy learning~\citep{srirama2024hrp, bahl2023affordances}, and latent action representations~\citep{ye2024latent, schmidt2024learning,bruce2024genie, kim2025uniskill, collins2025amplify}. 

Despite recent progress, existing approaches tend to fall into two extremes. Video generation methods~\citep{du2023learning, ko2023learning, zhou2024robodreamer, luo2025grounding} adopt a two-stage pipeline that first predicts future videos and then translates them into actions using an inverse dynamics model. While this paradigm effectively leverages rich visual and temporal priors, it introduces compounding errors during action prediction and incurs substantial latency due to the cost of generating high-fidelity videos.
In contrast, latent action models~\citep{ye2024latent, schmidt2024learning} predict the next frame in a learned latent space, yielding a compact representation that is easier to decode into actions. However, prior work typically predicts only a single latent action step, limiting the temporal abstraction that videos inherently provide.

Learning temporal abstractions (\emph{skills}) has been shown to substantially improve policy learning. In imitation learning, temporal abstractions help capture the non-Markovian structure present in offline demonstrations~\citep{pastor2009learning, nasiriany2022sailor, zhao2023aloha, zheng2024prise, mete2024quest}. Unsupervised skill discovery~\citep{eysenbach2018diayn, laskin2022cic, park2022lipschitz, park2023caw} enables agents to acquire diverse behaviors during pretraining without extrinsic rewards, while skill-based RL~\citep{ajay2020opal, pertsch2021accelerating, shi2022skillbasedrl, wilcoxson2025supe} leverages these learned skills for downstream policy learning, demonstrating strong performance on long-horizon tasks with sparse rewards. However, these works focus on state-based inputs, where skill learning is comparatively straightforward and does not address the complex dynamics inherent in videos.

A central question we explore is: \emph{How can we learn a good latent skill representation from videos for policy learning}? Recent cross-embodiment methods map human videos to robot actions by learning skills from pixel space~\citep{xu2023xskill, kim2025uniskill}. However, reconstruction objectives applied directly in pixel space often cause models to capture low-level visual details that are not useful for downstream semantic tasks~\citep{zheng2025rae}. Motivated by this, we propose to learn in an \emph{intermediate representation} that is more closely aligned with actions. Intermediate representations like optical flow and track points have been explored in policy learning~\citep{wen2023any, xu2024flow, gao2025flip}, but they are typically used directly for action inference~\citep{xu2024flow} or as auxiliary guidance~\citep{wen2023any, gao2025flip}, rather than as a basis for learning a compact skill latent space.

\begin{wrapfigure}[27]{r}{0.5\textwidth}
  \centering
  \includegraphics[width=\linewidth, trim=0 20 0 20, clip]{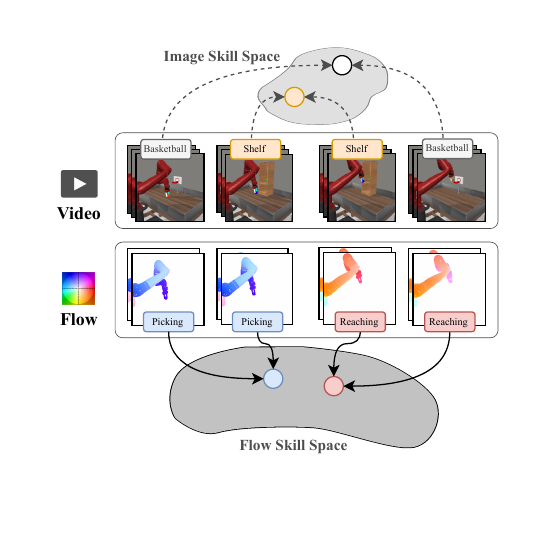}
  \vspace{-1.2cm}
      \caption{
  \textbf{Extracting skills from videos.} Videos contain composable \emph{skills} that appear across different tasks and scenes. Learning and planning in a skill space enables efficient multi-task learning and long-horizon planning. Learning skills from raw images (\emph{top}) often overfits to visual appearance. Instead, we learn skills from optical flow (\emph{bottom}), which captures motion patterns and better reflects the underlying actions.
  }
  \label{fig:overview}
\end{wrapfigure}

  

In this paper, we introduce \FullMethod{} (\Method{}), a framework for learning latent skills from optical flow sequences. Our framework consists of three components. First, we learn a latent skill space from optical flow extracted from action-free videos. Next, we train a skill policy that predicts skill tokens conditioned on the current observation. Finally, we train a lightweight module that maps predicted flows back to actions. This design provides two key benefits: (1) learning skills in an intermediate representation yields temporally coherent abstractions that improve high-level planning, and (2) the intermediate representation is easier to translate into low-level actions, mitigating compounding errors. Experiments on multi-task, long-horizon, and cross-embodiment settings demonstrate the effectiveness of our skill abstractions compared to prior video models, latent action models, and pixel-space baselines.

We summarize our contributions as follows:
(1) We propose \Method{}, a framework that learns skills from diverse, action-free videos by leveraging optical flow as an abstract representation.
(2) Our experiments show that \Method{} effectively leverages skill abstraction and achieves strong performance on long-horizon tasks, opening new directions for skill discovery from action-free visual data.

\section{Related work}
\label{sec:related-work}

\noindent\textbf{Learning robot policy from videos.} 
A growing body of work aims to leverage videos for robot learning without relying on expensive action-labeled datasets. Earlier work focuses on learning general visual representations and reward functions from videos~\citep{grauman2022ego4d, nair2022r3m, ma2023liv, zeng2024learning, hung2025victor}, which can then be used for downstream policy learning and reinforcement learning. Another line of research trains video generative models that predict future plans and subsequently convert them into actions via an inverse-dynamics module~\citep{du2023learning, mendonca2023structured, ajay2023compositional, ko2023learning, yang2024learning, videoworldsimulators2024, zhou2024robodreamer, liang2024dreamitate, luo2025grounding}. Our approach differs from this line of work in that we aim to learn directly in a skill latent space without generating video frames. A separate line of work learn latent actions~\citep{ye2024latent, schmidt2024learning, bruce2024genie, kim2025uniskill} from videos. These methods typically learn latent actions between two frames, rather than capturing a sequence of actions. AMPLIFY~\citep{collins2025amplify} is a concurrent work on arXiv that explores similar ideas. A key difference is that we use skills to directly produce actions, similar to the options framework, whereas they use skills primarily as guidance for action generation.

\noindent\textbf{Learning from intermediate representations.}
Beyond raw image observations, prior works have explored intermediate representations such as flow~\citep{ko2023learning, xu2024flow, gao2025flip}, keypoints~\citep{wen2023any, bharadhwaj2024track2act}, and affordances~\citep{bahl2023affordances, srirama2024hrp} for robotic manipulation. These approaches primarily employ such representations as auxiliary signals to guide policy learning. In contrast, our method leverages optical flow to learn a skill latent space, where this explicit use of flow to acquire reusable motion primitives provides a more scalable framework for long-horizon robot planning.

\noindent\textbf{Skills for decision making.}
To tackle long-horizon tasks, skill-based reinforcement learning (RL)~\citep{sutton1999between,  schaal2006dynamic, pastor2009learning, hausman2018learning, nasiriany2022augmenting, zhang2022efficient, zhang2024sprint} introduces temporal abstraction by representing policies as compositions of high-level skills or options~\citep{sutton1999between}. 
A large body of prior work aims to discover such skills in an unsupervised manner to accelerate downstream tasks learning, typically using heuristics or contrastive learning to extract skills from offline data~\citep{ajay2020opal, pertsch2021accelerating, nasiriany2022sailor, shi2022skillbasedrl, laskin2022cic}. 
Some approaches model low-level skills as discrete latent codes ~\citep{mete2024quest, zheng2024prise}. However, the aforementioned approaches require large amounts of action-labeled demonstration to learn meaningful skill representations.
While most prior works rely on discovering skills with robot demonstrations, some recent works have explored learning skill representations from action-free videos ~\citep{zhu2022bottom, tomar2023video, xu2023xskill}, reducing the need for ground-truth state or action labels. However, these methods often assume access to structured video data, such as paired human-robot demonstrations or predefined skill boundaries. In contrast, our approach learns skill representations directly from raw videos without task supervision. By leveraging optical flow as a proxy for actions, our framework enables the discovery of temporally abstract, composable skills across diverse tasks and environments, offering a more scalable and general approach to skill-based decision making. 
\section{Preliminary}
\label{sec:Preliminary}
\subsection{Problem setting}

We consider an action-free video dataset denoted as $\mathcal{D}_{\text{video}} = \{ (\boldsymbol{v}_i, \ell_i) \}_{i=1}^M$, consisting of $M$ video-language pairs. Each video $\boldsymbol{v}_i = (x_1, \ldots, x_T)$ is a sequence of RGB frames $x_t \in \mathbb{R}^{H \times W \times 3}$, and each corresponding language annotation $\ell_i$
is a natural language description of the task depicted in the video. In addition, we assume access to an action-labeled dataset, $\mathcal{D}_{\text{act}}$, which may either be a small subset of $\mathcal{D}_{\text{video}}$ with annotated actions or a larger dataset of interaction trajectories collected in the environment (e.g., play data). Notably, this dataset does not necessarily include language annotations.

\subsection{Finite Scalar Quantization}
Finite Scalar Quantization (FSQ)~\citep{mentzer2024finite} is a simple drop-in replacement for VQ-VAE that removes vector quantization to alleviate the optimization difficulties of VQ-VAE, such as representation collapse and the under-utilization of codewords. It replaces vector quantization with fixed scalar quantization, where continuous encoder outputs are directly discretized into a finite set of values, forming an implicit codebook without explicitly learning code vectors. This design eliminates the need for auxiliary losses and tricks required to maintain an explicit codebook.
\section{Methodology}
\label{sec:Methodology}

\begin{figure}
    \centering
    \includegraphics[width=0.90\linewidth]{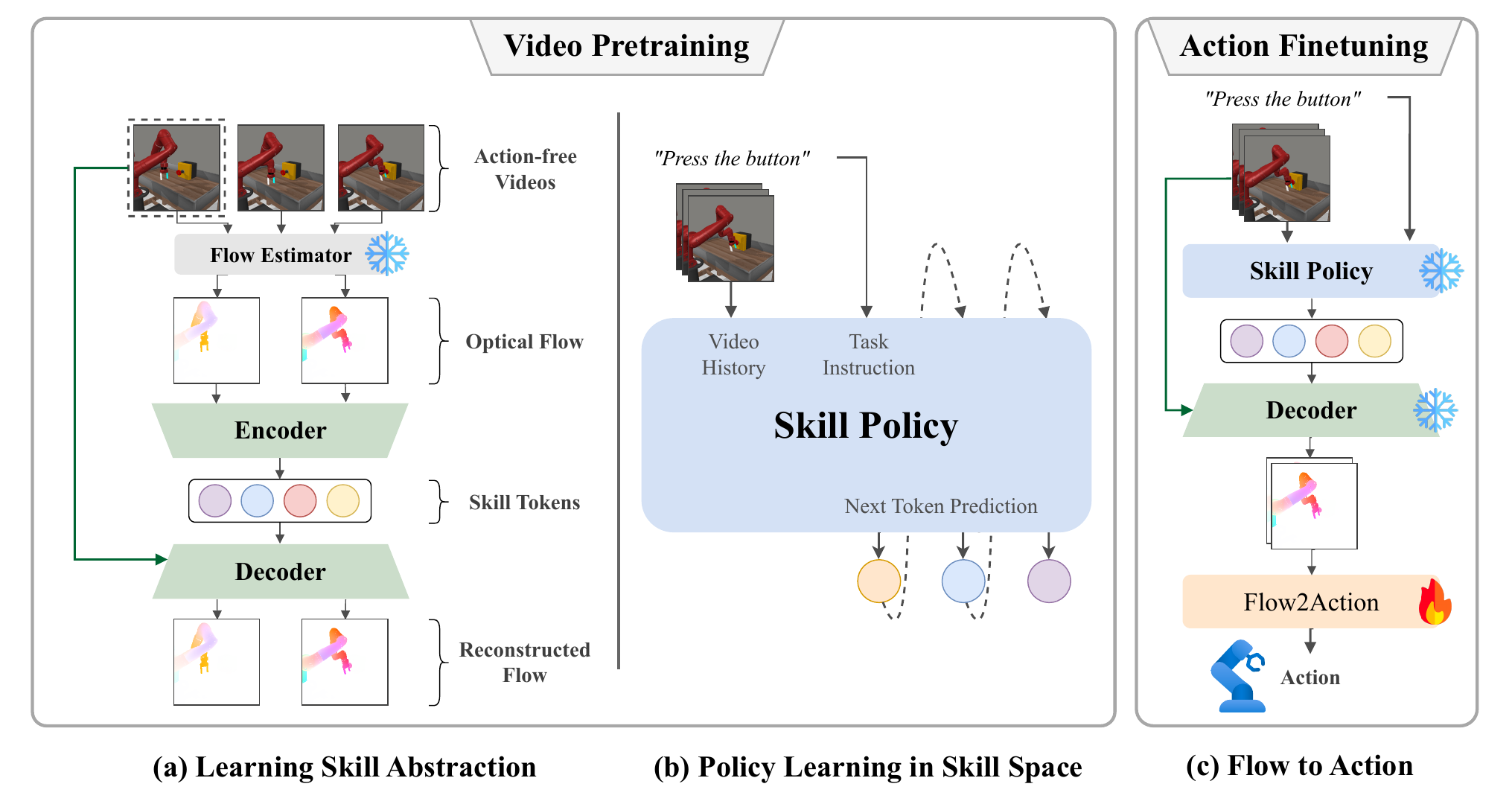}
    \caption{
    \textbf{\FullMethod{} (SOF).}
    (a) Learn an action abstraction from optical flow using latent variable models to capture motion patterns across tasks.
    (b) Learn a skill predictor to perform policy learning in the skill space. (c) Given the first frame and an instruction, \Method{} generates a skill plan, decodes it into optical flow using a decoder, and infers actions using a lightweight Flow2Action module. The Flow2Action module can be either learned or calculated.
    }
    \label{fig:framework}
\end{figure}

In this section, we present the key ideas and implementation details of \Method{}, Skill Abstraction from Optical Flow, as shown in Figure ~\ref{fig:framework}. Section~\ref{method:skill} describes how we learn action abstractions that predict a sequence of skill tokens from a sequence of action-free videos. In Section~\ref{method:decision_skill}, we introduce a skill policy that predicts these skill tokens based on video history and a given instruction, enabling policy learning in the skill space. The predicted skills are then decoded into optical flow conditioned on the current image. Finally, Section~\ref{method:flow2act} outlines two approaches for inferring action sequences from the predicted optical flow: a learning-based method and a learning-free alternative.

\subsection{Learning skill abstractions with optical flow} \label{method:skill}

To enable the learning of \emph{composable skill abstractions} from video datasets $\mathcal{D}_{\text{video}}$ composed of compound behaviors, our first step is to discover and aggregate recurring motion patterns into reusable skills across diverse demonstrations. However, with only videos, we lack both action labels and environment state transitions, making it challenging to infer actionable structure. To address this, we propose using optical flow as a surrogate for action labels. Optical flow offers several advantages: it is \emph{action-directed}, capturing relative pixel movement that results from robot actions, and \emph{noise-resistant}, as it focuses only on changes between consecutive frames while ignoring background and other motion-irrelevant noise.

Formally, given a video $\boldsymbol{v}_i = (x_1, \dots, x_T)$, we compute optical flow between each consecutive frame using an off-the-shelf estimator, yielding a flow sequence $\boldsymbol{\delta}_{i} = {\delta_1, \dots, \delta_{T-1} } $. In our implementation, we use FlowFormer++~\citep{shi2023flowformer++} for real-world videos and NeuFlow-v2~\citep{zhang2024neuflow} for simulated environments.

With the extracted flow sequences $\boldsymbol{\delta}_i$, we learns discrete skill abstractions from action sequences using a quantized autoencoder. In our setting, the encoder $\phi_\theta$ processes a flow segment $\delta_{t:t+H-1}$ of length $H$ and encodes them into a sequence of discrete latent skill tokens $\boldsymbol{c} = ( c_1, \cdots, c_n)$ via FSQ:
\begin{align}
\mathbf{c} = \text{FSQ} \left( \phi_\theta \left( \delta_{t:t+H-1} \right) \right)
\end{align}
The decoder $\psi_\theta$ reconstructs the original flow segment from these tokens. Crucially, we incorporate a positional inductive bias by conditioning the decoder on the initial frame $x_t$, leveraging the fact that optical flow captures relative motion. This conditioning enables the model to disentangle skill-relevant dynamics from absolute position, which may vary across demonstrations but is irrelevant to the underlying motion primitive.

The autoencoder is trained using a flow reconstruction loss:
\begin{align}
\mathcal{L}_{\text{recon}}(\theta) = \left\| \psi_{\theta}(\text{FSQ}(\phi_{\theta}(\delta_{t:t+H-1})), x_{t}) - \delta_{t:t+H-1} \right\|_1.
\end{align}

\subsection{Learning decision making with skills} \label{method:decision_skill}

After learning skill abstractions, we train a skill policy $\pi_\omega(c_t \mid x_t, e)$ to predict the skill based on the current frame $x_t$ and the task embedding $e$. The image observation is encoded using a learned vision encoder, which is trained jointly with the skill policy in an end-to-end manner. Notably, the model is also trained using the action-free video dataset $\mathcal{D}_{\text{video}}$. 

Unlike prior methods such as ATM~\citep{wen2023any}, which rely on multiple synchronized camera views, or QueST~\citep{mete2024quest}, which requires privileged state information, our method uses only third-person visual observations—similar to learning-from-video approaches~\citep{du2023learning, ko2023learning, ye2024latent}—making it better suited for unstructured, real-world video data.

To model the temporal dependencies between skill tokens, we follow QueST~\citep{mete2024quest} and employ a decoder-only Transformer as the skill predictor. The model autoregressively generates the skill sequence based on the current image and task context:
\begin{align}
    \pi_\omega(c_{1:n} \mid x_t, e) = \prod_{i=1}^{n} \pi_\omega(c_i \mid c_{<i}, x_t, e)
\end{align}
The skill policy is optimized using the negative log-likelihood objective:
\begin{align}
    \mathcal{L}_{\text{skill}}(\omega) = - \sum_{i=1}^{n} \log \pi_{\omega}(c_i \mid c_{<i}, x_t, e)
\end{align}

\subsection{Action execution via predicted optical flow plan} \label{method:flow2act}

With the modules described in Section~\ref{method:skill} and Section~\ref{method:decision_skill}, our model predicts future optical flows that represent the actions a robot should take to complete a given task, based on the history of image observations and the task instruction. In this section, we introduce two approaches—one learning-free and one learning-based—to convert the predicted flows into executable actions.

\noindent\textbf{Learning-free.}
For the learning-free method, we adopt the action regression technique from AVDC~\citep{ko2023learning}, which infers actions directly from optical flow. Specifically, it estimates rigid transformations of the target object from the optical flow sequence, producing a series of SE(3) transformations. These transformations are then executed using a heuristic grasping strategy combined with a simple path-following policy to achieve the object transformation.

\noindent\textbf{Learning-based.}
However, AVDC relies on several assumptions about the environment, such as the availability of depth information and accurate segmentation masks, which may not be accessible or reliable in real-world scenarios. To overcome these limitations, we propose a learning-based alternative that frames flow-to-action mapping as a regression problem. We fine-tune a lightweight flow-to-action model using a small set of videos with ground-truth action labels, enabling the model to infer actions from flow in a data-efficient and environment-agnostic way. In our experiments, we show that this approach performs well even with limited labeled data, demonstrating strong potential for action inference compared to traditional methods such as inverse dynamics models~\citep{agrawal2016learning}.

\section{Experiments}
\label{sec:Experiments}
Our experiments seek to answer the following questions: (1) Can \Method{} efficiently acquire diverse skills from multi-task video data? (2) Do the acquired skills enhance performance on long-horizon tasks? (3) Can the learned skills generalize across different embodiments?

\begin{figure}[H]
  \centering
  \begin{subfigure}[b]{0.32\linewidth}
    \includegraphics[clip,width=\textwidth]{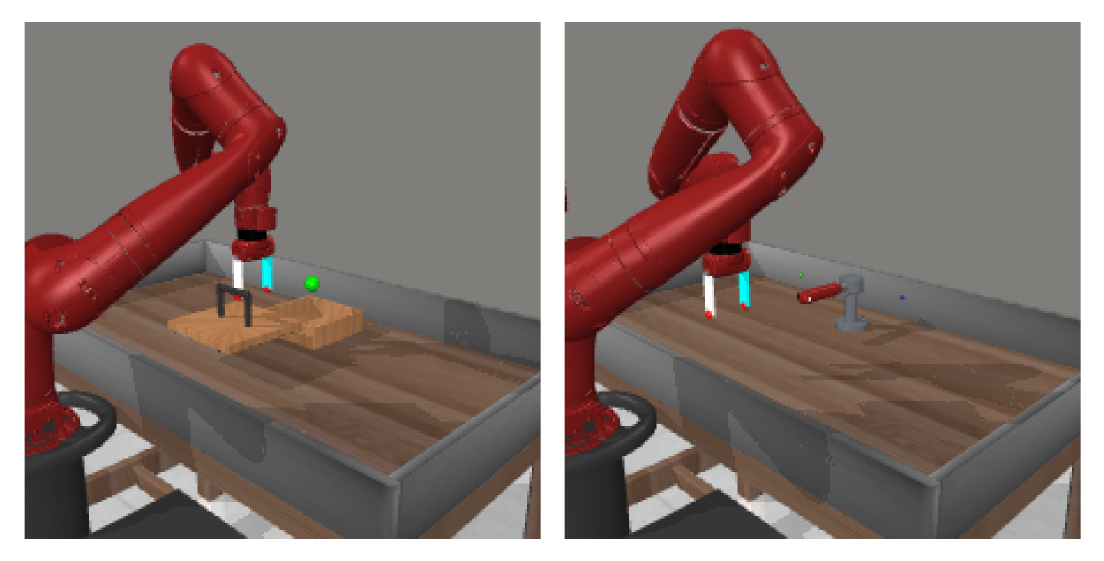}
    \caption{MetaWorld}
    \label{fig:env_mw}
  \end{subfigure}
  \hfill
  \begin{subfigure}[b]{0.32\linewidth}
    \includegraphics[clip,width=\textwidth]{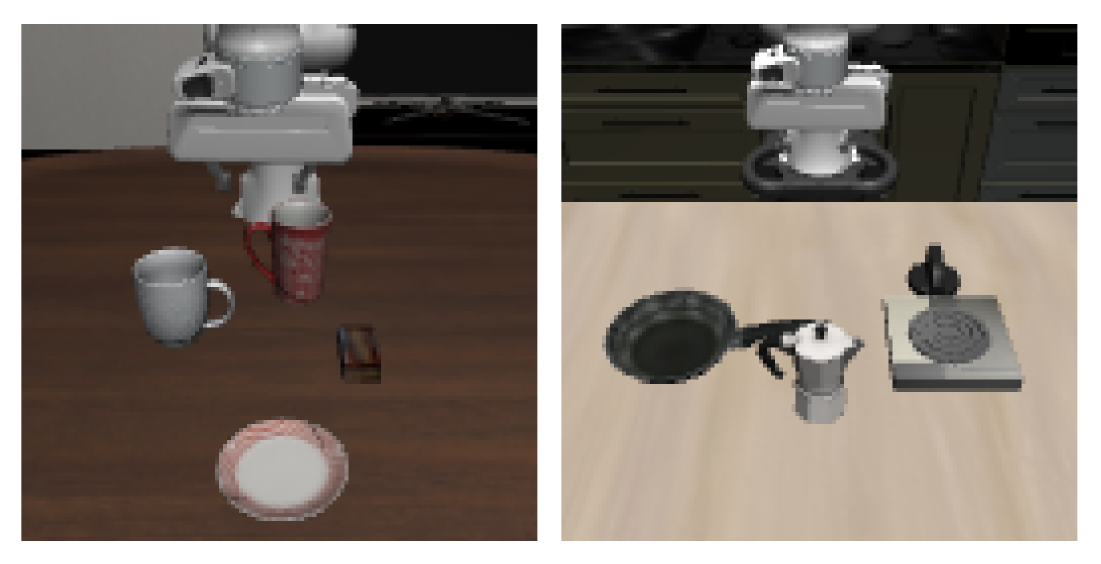}
    \caption{LIBERO}
    \label{fig:env_libero}
  \end{subfigure}
  \hfill
  \begin{subfigure}[b]{0.32\linewidth}
    \includegraphics[clip,width=\textwidth]{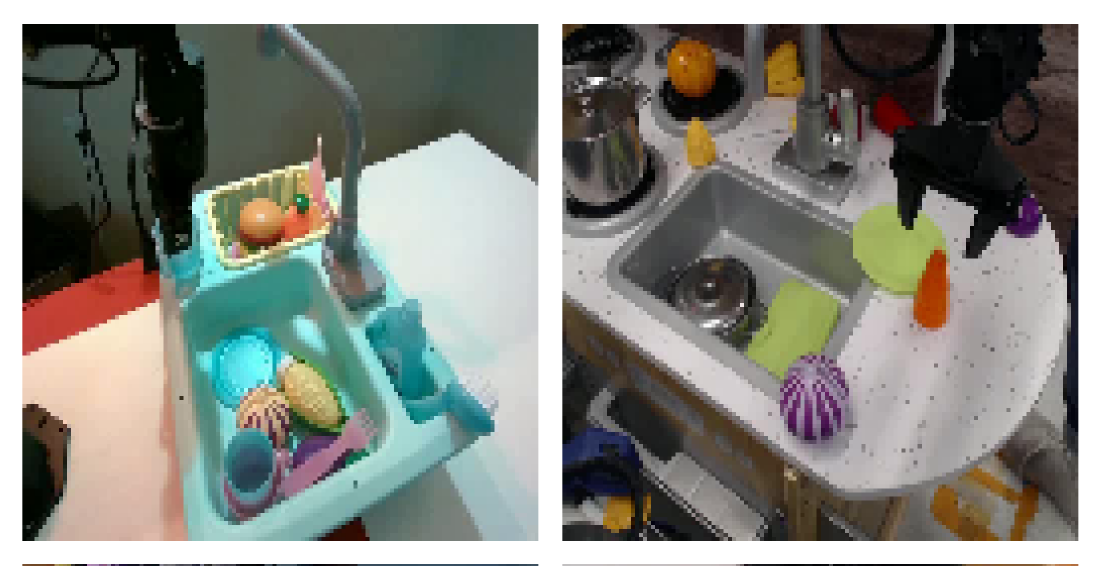}
    \caption{BridgeData V2}
    \label{fig:env_bridge}
  \end{subfigure}
  \caption{\textbf{Environmental setups}: (a) \textbf{MetaWorld} is a simulation benchmark featuring a variety of manipulation tasks. We used it to evaluate multi-task performance and cross-embodiment generalization.
(b) \textbf{LIBERO} is a simulation benchmark for lifelong robot learning. We use it to study multi-task and long-horizon performance.
(c) \textbf{BridgeData V2} is a real-world dataset of manipulation behaviors. We use it to evaluate our skills on diverse environments and tasks.}
  \label{fig:env}
\end{figure}



\subsection{Baselines}

\label{sec:baseline}

\noindent\textbf{Behavior Cloning (BC).} We implement multi-task BC. Specifically, we concatenate image features with task instruction features encoded by CLIP, and feed the combined representation into a 3-layer MLP to predict actions. The model is trained using mean squared error loss.

\noindent\textbf{Diffusion Policy (DP; \citealp{chi2023diffusionpolicy})} is a state-of-the-art imitation learning algorithm that employs a conditional diffusion model as a policy, allowing for modeling multimodal action distributions. We adopt the CNN-based Diffusion Policy, which uses a 1D convolutional U-Net to denoise action sequences sampled from a Gaussian prior, conditioned on RGB observations.

\noindent\textbf{UniPi}~\citep{du2023learning} trains a text-conditioned video diffusion model to generate video plans using action-free video data $\mathcal{D}_{\text{video}}$, as in our setup. An inverse dynamics model, trained separately on action-labeled data $\mathcal{D}_{\text{act}}$, is then used to infer actions from the generated videos. To mitigate error accumulation during open-loop execution, we apply a replanning strategy~\citep{ko2023learning} that regenerates video plans when the robot's behavior deviates from the original trajectory.

\noindent\textbf{AVDC}~\citep{ko2023learning} uses a text-conditioned video diffusion model with a learning-free approach that infers actions from optical flow, depth, and object masks.

\noindent\textbf{LAPA}~\citep{ye2024latent} is a vision-language-action model that learns from videos. It consists of a latent action pretraining stage on action-free video data $\mathcal{D}_{\text{video}}$, where the goal is to learn latent actions that can predict future frames. This is followed by an action finetuning stage on action-labeled data $\mathcal{D}_{\text{act}}$. Following the original setup, we use the 7B LWM-Chat-1M~\citep{liu2025world} as the base VLM.

\subsection{Environments}

\label{sec:env}

We evaluate on MetaWorld and LIBERO benchmarks. For multi-task settings, we use 9 MetaWorld tasks (third-person camera) and 10 LIBERO-GOAL tasks (front-facing camera). Each task includes 50 action-free videos ($\mathcal{D}_{\text{video}}$) and 10 action-labeled trajectories ($\mathcal{D}_{\text{act}}$), used for fine-tuning (LAPA) and Flow2Action (Ours). For long-horizon tasks, we select 4 from LIBERO-10, as most baselines fail on the rest with video-only supervision. Each has 50 action-free videos and 10 action-labeled trajectories. For cross-embodiment evaluation, we use MetaWorld to test generalization between Sawyer and Panda. See Section~\ref{sec:cross_embod} for details. We also conduct a skill space analysis on BridgeData V2~\citep{walke2023bridgedata} to test real-world skill abstractions. We learn a latent variable model for skill representation and train a skill predictor to infer flow plans.

\begin{table}[t]
\centering
\caption{
\textbf{Multi-Task Learning on MetaWorld.} 
\Method{} effectively utilizes action-free video data and outperforms all baselines. We compare against (1) multi-task BC baselines trained on action-labeled data $\mathcal{D}_{\text{act}}$ (10 demos per task), and (2) video-based methods trained on 50 demos per task that fine-tune or learn a module (e.g., IDM, Flow2Action) using $\mathcal{D}_{\text{act}}$. The highest score is highlighted in \textbf{bold}, and the second-highest score is \underline{underlined}.
}
\vspace{-0.2cm}
{
\setlength{\tabcolsep}{3pt}
\scalebox{0.75}{%
\begin{tabular}{l c c c c c}
\toprule
& door-open & door-close & bin-picking & box-close & drawer-open   \\
\midrule
BC &\numerr{0.64}{0.06} & \bf \numerr{1.00}{0.00} & \numerr{0.00}{0.00} & \bf \numerr{0.20}{0.07} & \numerr{0.63}{0.02} \\
DP &  \numerr{0.00}{0.00} & \numerr{0.84}{0.05} & \numerr{0.00}{0.00} & \numerr{0.00}{0.00} & \bf \numerr{1.00}{0.00} \\
AVDC & \numerr{\underline{0.84}}{0.04} & \numerr{\underline{0.92}}{0.04} & \numerr{0.00}{0.00} & \numerr{0.04}{0.00} & \numerr{0.02}{0.02}  
  \\
LAPA & \numerr{0.00}{0.00} & \numerr{0.00}{0.00} & \numerr{0.00}{0.00} & \numerr{0.00}{0.00} & \numerr{0.00}{0.00} \\

\midrule
\Method{} (Ours) & \bf  \numerr{0.98}{0.03} & \bf \numerr{1.00}{0.00} &\bf  \numerr{0.24}{0.07} & \numerr{\underline{0.12}}{0.07} &  \numerr{\underline{0.78}}{0.04} \\

\midrule
\midrule




& faucet-close & faucet-open & handle-press & assembly & \textbf{Overall} \\
\midrule

BC &  \bf \numerr{0.78}{0.04} & \bf \numerr{1.00}{0.00} & \bf \numerr{0.87}{0.03} & \numerr{0.00}{0.00} & \numerr{\underline{0.57}}{0.01}\\
DP & \numerr{0.06}{0.02} & \numerr{0.86}{0.07} & \numerr{0.00}{0.00} & \numerr{0.00}{0.00} & \numerr{0.31}{0.01}  \\ 
AVDC & \numerr{0.24}{0.04} & \numerr{0.78}{0.02} & \numerr{\underline{0.72}}{0.04} & \numerr{0.00}{0.00} & \numerr{0.42}{0.02} \\
LAPA &\numerr{0.17}{0.08} & \numerr{0.28}{0.11} & \numerr{0.65}{0.11} & \numerr{\underline{0.12}}{0.04} & \numerr{0.14}{0.02} \\
\midrule
\Method{} (Ours) &\numerr{\underline{0.62}}{0.06} &  
 \numerr{\underline{0.99}}{0.02} & \numerr{0.69}{0.06} & \bf  \numerr{0.82}{0.07} & \bf \numerr{0.69}{0.02} \\

\bottomrule
\end{tabular}
}
}

\vspace{5pt}
\label{tab:mw-main}
\vspace{-3pt}
\end{table}










\begin{table}[t]
\centering
\caption{\textbf{Multi-Task Learning on LIBERO-GOAL.} 
\Method{} clearly outperforms the baselines on repetitive tasks -- primarily those involving picking up objects and placing them elsewhere -- highlighting the advantages of reusable skills. However, \Method{} underperforms compared to BC and DP on tasks that involve handling small objects (e.g., bottle) or require distinct motions (e.g., open a drawer).
}
\vspace{-0.2cm}
{
\setlength{\tabcolsep}{3pt}
\scalebox{0.75}{%
\begin{tabular}{l c c c c c c}
\toprule 
& put-bowl-stove & put-bowl-cabinet & push-plate-stove & put-bottle-cabinet & put-cream-bowl    \\
\midrule
BC & \numerr{\underline{0.38}}{0.23} & \numerr{\underline{0.06}}{0.01} & \numerr{\underline{0.31}}{0.10} & \bf \numerr{0.39}{0.05} & \bf \numerr{0.16}{0.06}  \\
DP & \numerr{0.10}{0.05} & \numerr{0.04}{0.04} & \numerr{0.18}{0.04} & \numerr{0.01}{0.01} & \numerr{0.02}{0.03} \\
LAPA &  \numerr{0.00}{0.00} & \numerr{0.00}{0.00} & \numerr{0.00}{0.00} & \numerr{0.00}{0.00} & \numerr{0.00}{0.00}\\

\midrule
\Method{} (Ours) & \bf \numerr{0.64}{0.06} & \bf \numerr{0.30}{0.06} & \bf \numerr{0.57}{0.06} & \numerr{\underline{0.01}}{0.01} & \numerr{\underline{0.07}}{0.02}  \\

\midrule\midrule

& turn-on-stove & put-bowl-plate & put-bottle-rack & open-middle-drawer &  open-top-drawer &  Avg. \\

\midrule
BC & \numerr{\underline{0.77}}{0.05} & \numerr{\underline{0.14}}{0.09} & \numerr{\underline{0.03}}{0.03} & \numerr{\underline{0.06}}{0.05} & \numerr{0.00}{0.00} & \numerr{\underline{0.23}}{0.06} \\
DP & \bf  \numerr{0.94}{0.04} & \numerr{0.01}{0.01} & \numerr{0.00}{0.00} & \bf \numerr{0.46}{0.13} & \numerr{0.00}{0.00} & \numerr{0.18}{0.01} \\
LAPA &  \numerr{0.00}{0.00} & \numerr{0.00}{0.00} & \numerr{0.00}{0.00} & \numerr{0.00}{0.00} & \numerr{0.00}{0.00} & \numerr{0.00}{0.00}\\
\midrule
\Method{} (Ours) & \numerr{0.71}{0.06} & \bf \numerr{0.21}{0.07} & \bf \numerr{0.03}{0.01} & \numerr{0.00}{0.00} & \numerr{0.00}{0.00} & \bf \numerr{0.25}{0.02} \\

\bottomrule
\end{tabular}
}
}
\label{tab:libero-main}
\vspace{-3pt}
\end{table}

\subsection{Multi-task learning}
\label{sec:multi_task}

We evaluate multi-task learning on the MetaWorld and LIBERO-GOAL benchmarks. On MetaWorld, \Method{} consistently outperforms both video-based baselines and multi-task BC baselines, using the same amount of action-labeled data. We found that LAPA, which learns latent actions from action-labeled data, fails to perform grasping tasks effectively. On LIBERO-GOAL, which comprises tasks involving similar motions (e.g., picking and placing objects), our method successfully captures reusable skills and outperforms the baselines. However, we observe two limitations: it struggles to detect small objects such as cream containers or bottles, and it fails to execute motions requiring large rotations, such as opening drawers. In the LIBERO environment, where scenes across tasks are visually similar and the gripper is not easily distinguishable, these challenges lead to frequent failures in image based method such as LAPA.

\subsection{Long-horizon tasks}
\label{sec:long_horizon}

We compare \Method{} against BC and DP on LIBERO-10 for long-horizon tasks. As shown in Table~\ref{tab:long_horizon}, when using 10 demonstrations across all methods, both BC and DP struggle with long-horizon tasks, while our method effectively leverages reusable skills, plans in the skill space, and successfully completes long-horizon tasks. When the number of demonstrations for BC and DP is increased, their performance becomes comparable to ours.

\begin{table}
  \caption{\textbf{Long-horizon on LIBERO-10.} Using 10 action-labeled demonstrations per task for BC, DP, and \Method{}, we observe that BC and DP struggle with long-horizon tasks, whereas \Method{} efficiently composes reusable skills to solve them. When the number of demonstrations for BC and DP is increased to 30, their performance becomes comparable to \Method{}.}
  \label{tab:long_horizon}
  \centering
  \resizebox{\columnwidth}{!}{
  \begin{tabular}{cccccc}
    \toprule
    & put-soup\_sauce-basket & turn-on-stove-put-moka-pot & put-mug-left-right & put-mug-left-pudding-right & Overall \\
    \midrule
    BC (10 demos) & \numerr{0.00}{0.00} & \numerr{0.09}{0.05} & \numerr{0.00}{0.00} & \numerr{0.02}{0.02} & \numerr{0.03}{0.01} \\
    BC (30 demos) & \numerr{0.00}{0.00} & \numerr{0.23}{0.10} & \numerr{0.03}{0.00} & \numerr{0.02}{0.02} & \numerr{0.07}{0.03}  \\
    DP (10 demos) & \numerr{0.00}{0.00} & \numerr{0.06}{0.04} & \numerr{0.00}{0.00} & \numerr{0.00}{0.00} & \numerr{0.01}{0.01} \\ 
DP (30 demos) & \numerr{0.02}{0.03} & \numerr{0.44}{0.06} & \numerr{0.09}{0.02} & \numerr{0.06}{0.04} & \numerr{0.15}{0.01} \\ 
UniPi  & \numerr{0.00}{0.00} & \numerr{0.00}{0.00} & \numerr{0.00}{0.00} & \numerr{0.00}{0.00} & \numerr{0.00}{0.00} \\ 
\midrule
\Method{} (Ours) & \numerr{0.08}{0.03} & \numerr{0.42}{0.06} & \numerr{0.06}{0.01} & \numerr{0.10}{0.03} & \numerr{0.16}{0.01}  \\
    \bottomrule
  \end{tabular}
  }
  \vspace{-15pt}
\end{table}

\subsection{Cross-embodiment transfer}
\label{sec:cross_embod}

\noindent\textbf{Transferring learned skills.}
We test whether learned skills generalize across robot embodiments using the Franka Panda and Sawyer arms. In stage one, videos from both arms are used to learn a shared skill space; in stage two, policy learning is trained on only one arm’s data. Flow2Action modules are trained separately per arm to map optical flow to low-level actions. We define “topline” as performance when training on all tasks per embodiment, serving as an upper bound.

\begin{figure}[h]
  \centering
  \begin{subfigure}[b]{0.47\linewidth}
    \includegraphics[clip,width=\textwidth]{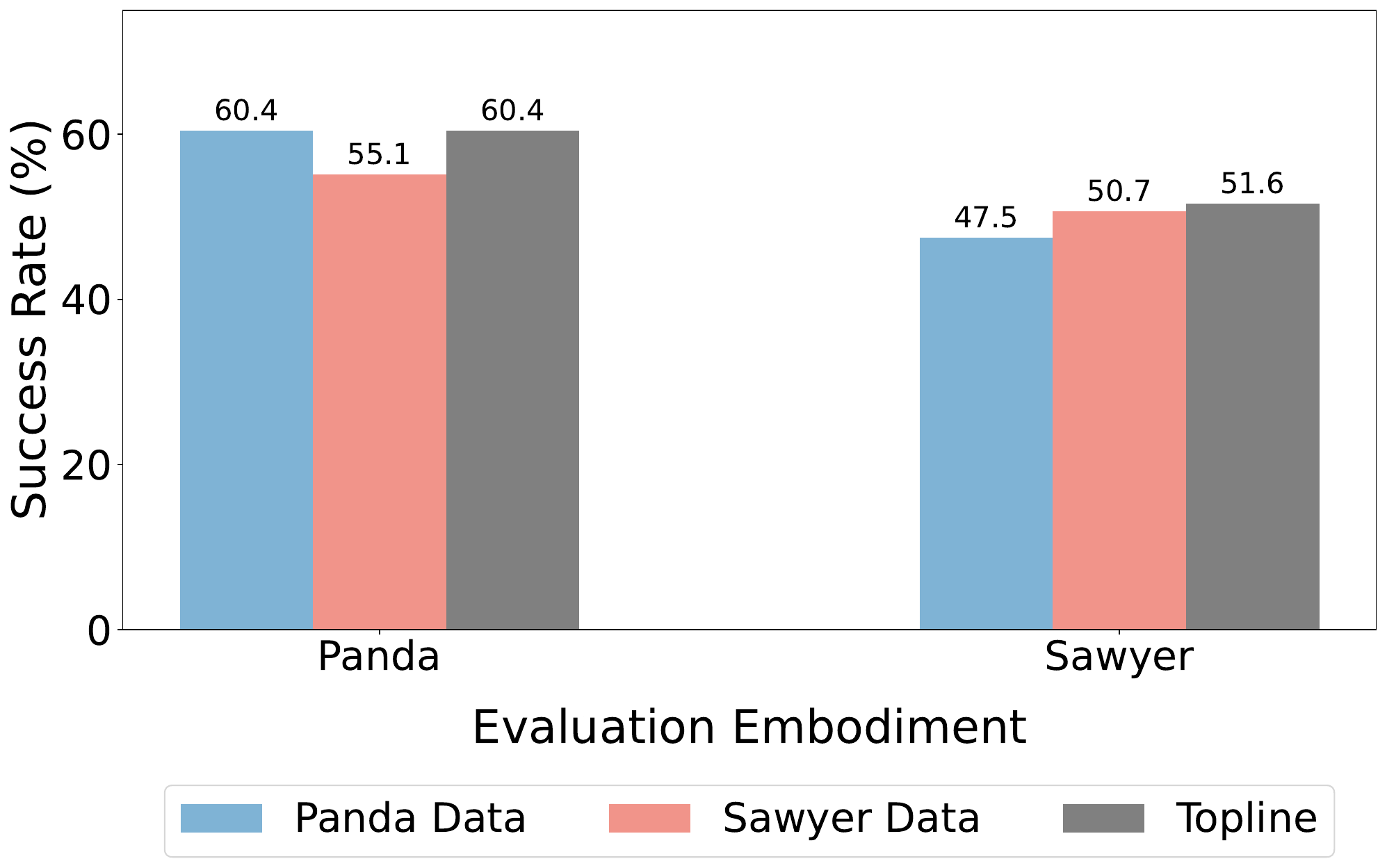}
    \caption{Skill Policy Transfer}
    \label{fig:cross_embod}
  \end{subfigure}
  \hfill
  \begin{subfigure}[b]{0.48\linewidth}
    \includegraphics[clip,width=\textwidth]{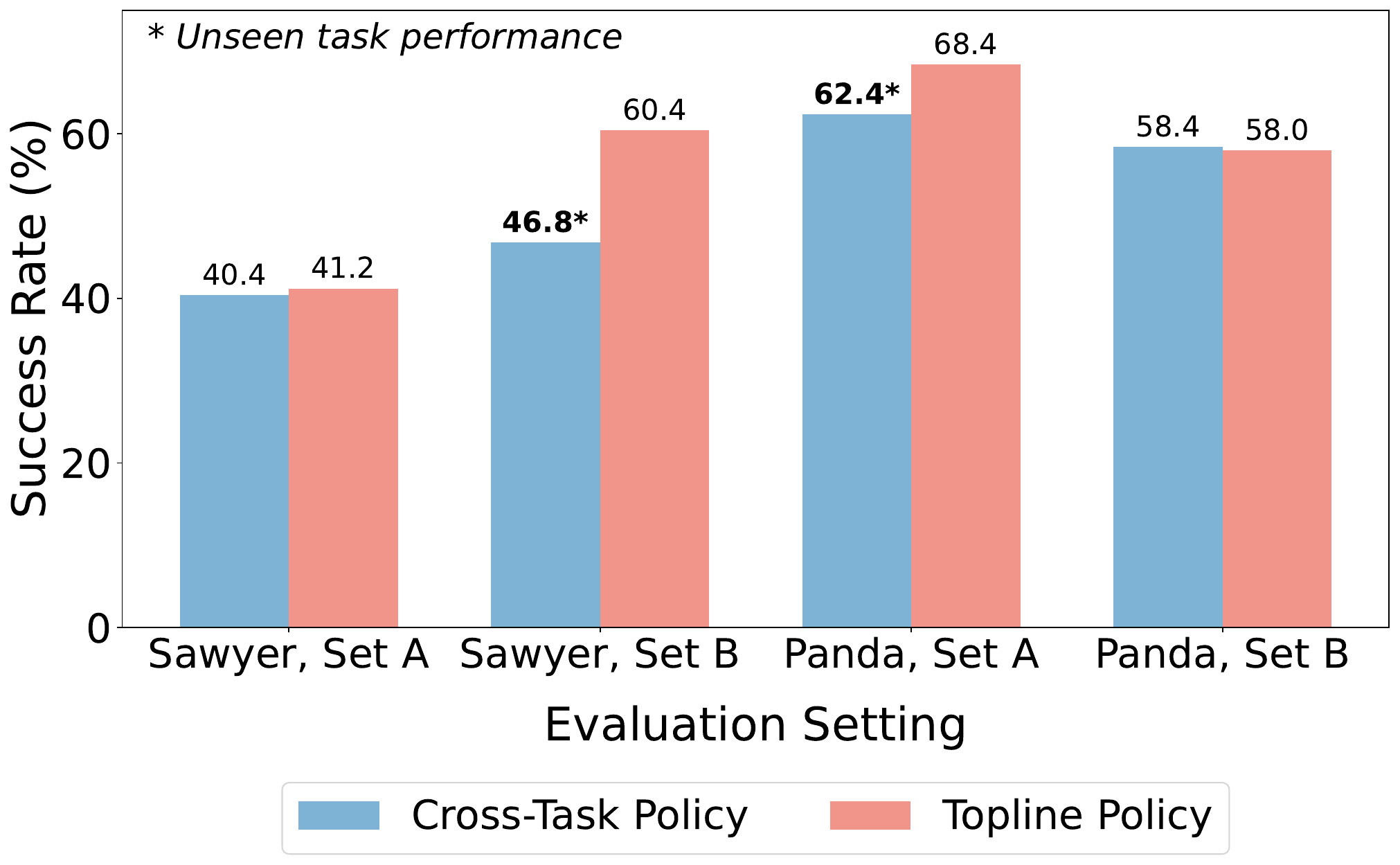}
    \caption{Cross-Task Transfer}
    \label{fig:cross_task}
  \end{subfigure}
  \caption{\textbf{Cross-embodiment transfer.} Average success rates on MetaWorld: 
        \textbf{(a) Skill policy transfer.} We train the skill abstraction with both Panda and Sawyer data. In the policy training stage, only one embodiment's data is used. \emph{Topline} shows results trained on all tasks per embodiment. The results show that shared skill representation enable effective transfer across embodiments.
        \textbf{(b) Cross-task transfer.}
        We partition tasks into disjoint sets, \sawyer{A} and \panda{B}. The \emph{cross-task policy} is trained on Sawyer data from \sawyer{A} and Panda data from \panda{B}, while the \emph{topline policy} is trained on the full dataset. The results indicate successful transfer, even when a task is unseen for one embodiment.}
  \label{fig:cross}
\end{figure}

\begin{figure}
  \centering
  \begin{subfigure}[b]{0.42\linewidth}
    \includegraphics[width=\linewidth]{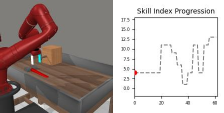}
    \caption{Sawyer}
    \label{fig:cross_sawyer}
  \end{subfigure}
  \hfill
  \begin{subfigure}[b]{0.42\linewidth}
    \includegraphics[width=\linewidth]{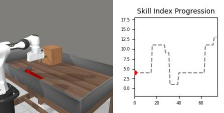}
    \caption{Panda}
    \label{fig:cross_panda}
  \end{subfigure}
  \caption{\textbf{Skill token index progress across embodiments.} Sawyer and Panda arms follow similar skill token sequences, with minor differences in skill duration.}
  \label{fig:cross_embodiment_skill_progress}
  \vspace{-20pt}
\end{figure}

As shown in Table~\ref{fig:cross_embod}, decision-making transfers well to unseen embodiments. Figure~\ref{fig:cross_embodiment_skill_progress} shows both arms follow similar skill-token sequences for the same task, with only minor timing differences due to embodiment-specific dynamics. This indicates the shared skill space captures consistent high-level behaviors while accommodating low-level variations.

\noindent\textbf{Transfer for visual perception and decision making on unseen tasks.}
We further test unseen task generalization by splitting 10 Metaworld tasks between the Panda and Sawyer, training both skill abstraction and decision making only on the assigned tasks (Appendix~\ref{appendix:cross-embodiment-tasks}).
The results in Table~\ref{fig:cross_task} indicate that, despite the absence of training data for certain tasks for a robot, \Method{} can still use skills from other embodiments to achieve high performance.

\subsection{Analysis}

\begin{figure}
  \centering
  \begin{subfigure}[b]{0.32\linewidth}
    \includegraphics[width=\linewidth]{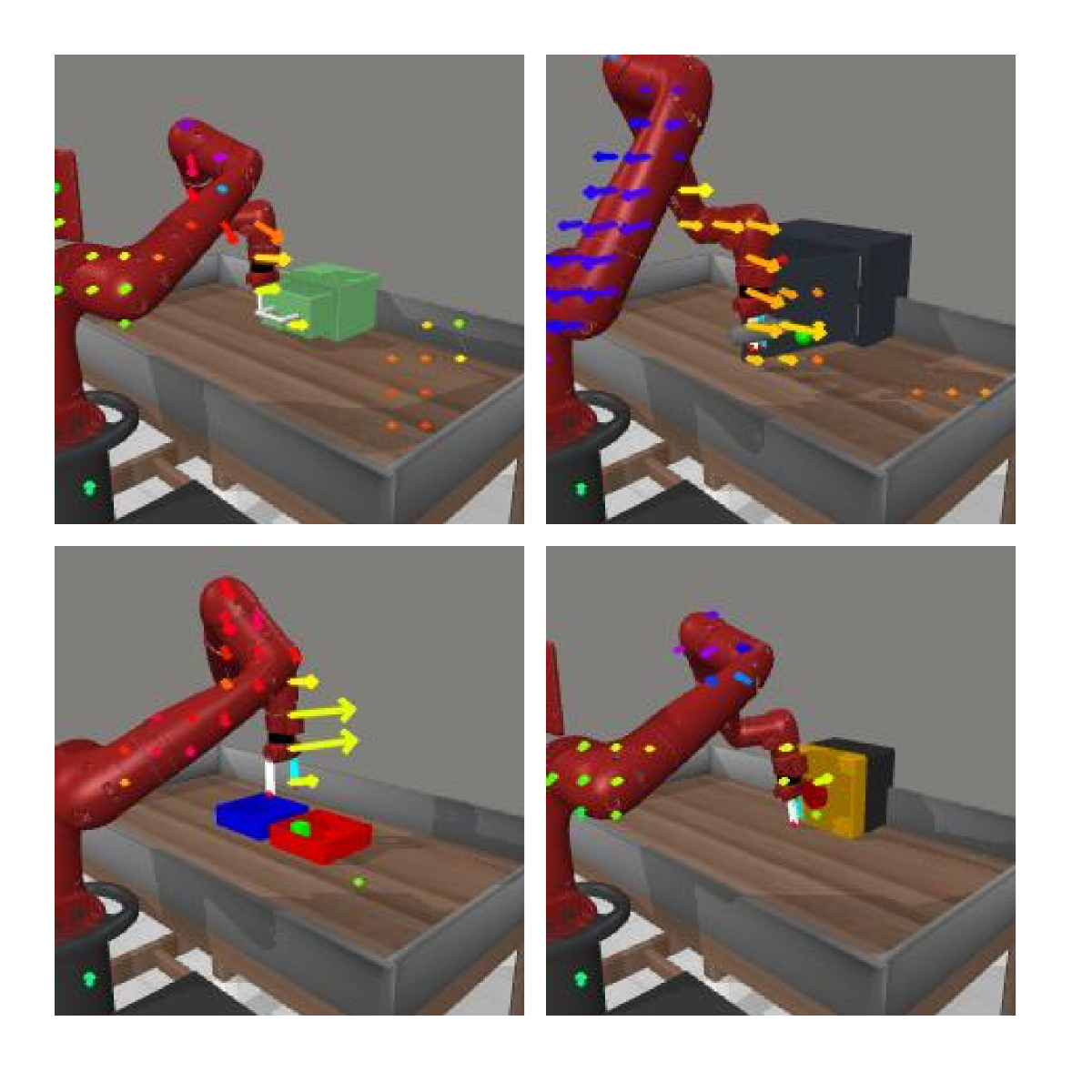}
    \caption{Different Tasks \& Scenes}
    \label{fig:mw_skill}
  \end{subfigure}
  \hfill
  \begin{subfigure}[b]{0.32\linewidth}
    \includegraphics[width=\linewidth]{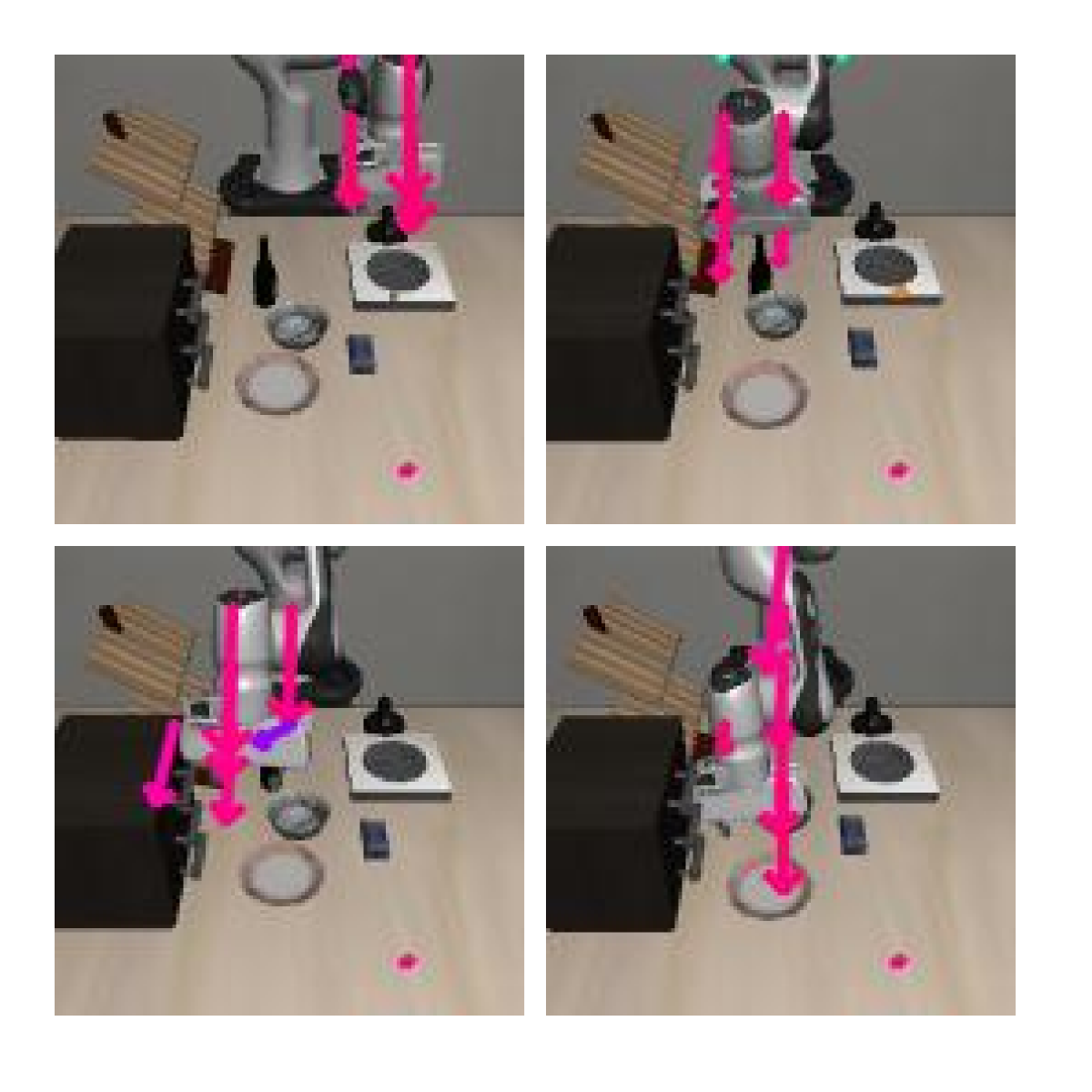}
    \caption{Different Objects \& Positions}
    \label{fig:lg_skill}
  \end{subfigure}
  \hfill
  \begin{subfigure}[b]{0.32\linewidth}
    \includegraphics[width=\linewidth]{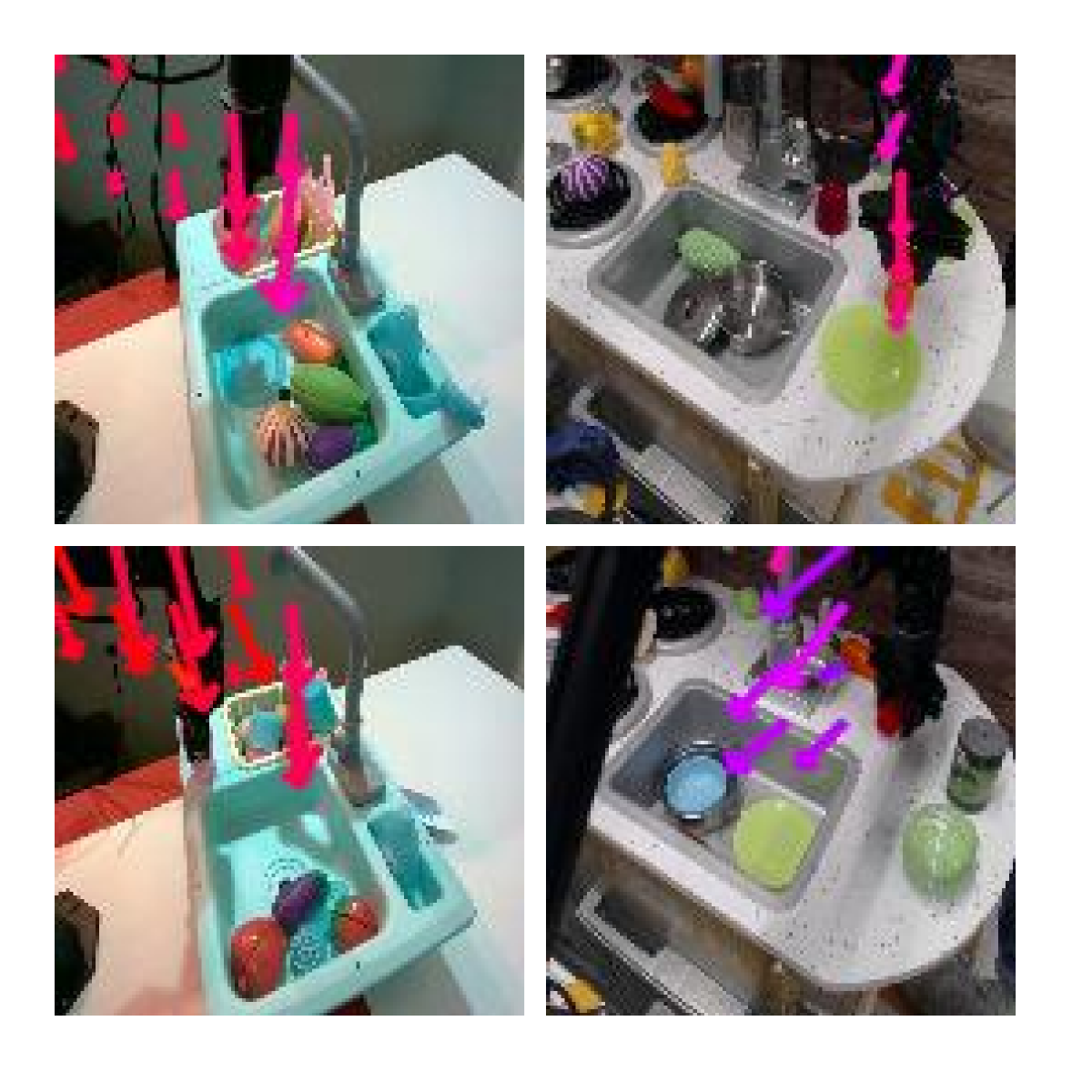}
    \caption{Real World}
    \label{fig:rw_skill}
  \end{subfigure}
  \caption{\textbf{Skill Token Analysis}: Each figure shows the optical flow plan that corresponds to \emph{the same skill token}. (a) Different tasks and scenes involving similar motion patterns are grouped together (b) Visually distinct objects in the same scene, positioned differently, are grouped together due to shared motion (c) Visually diverse real-world scenes are grouped together by shared motion patterns. }
  \label{fig:skill_token}
  \vspace{-8pt}
\end{figure}

\noindent\textbf{Skill token analysis in multi-task setting.} 
We analyze learned skills in a multi-task setting to test whether similar motions map to the same skill across conditions. To handle the large codebook (1024), we cluster embeddings into 16 groups with K-means, then sample and visualize skills using the current frame and predicted optical flow for the next $k$ steps. The flow plan is illustrated with arrows: different colors indicate different directions, while color intensity reflects motion magnitude.

Figure~\ref{fig:skill_token} shows that tokens generalize across varied objects, tasks, and layouts: pushing and grasping share the same token (Fig.~\ref{fig:mw_skill}), spatial invariance holds across positions and objects (Fig.~\ref{fig:lg_skill}), and Bridge results (Fig.~\ref{fig:rw_skill}) demonstrate consistent motion clustering despite real-world variability. These results highlight robust skill abstraction across simulation and real settings.

\begin{figure}[t]
\centering

\begin{minipage}[c]{0.45\linewidth}
  \centering
  \includegraphics[width=\linewidth]{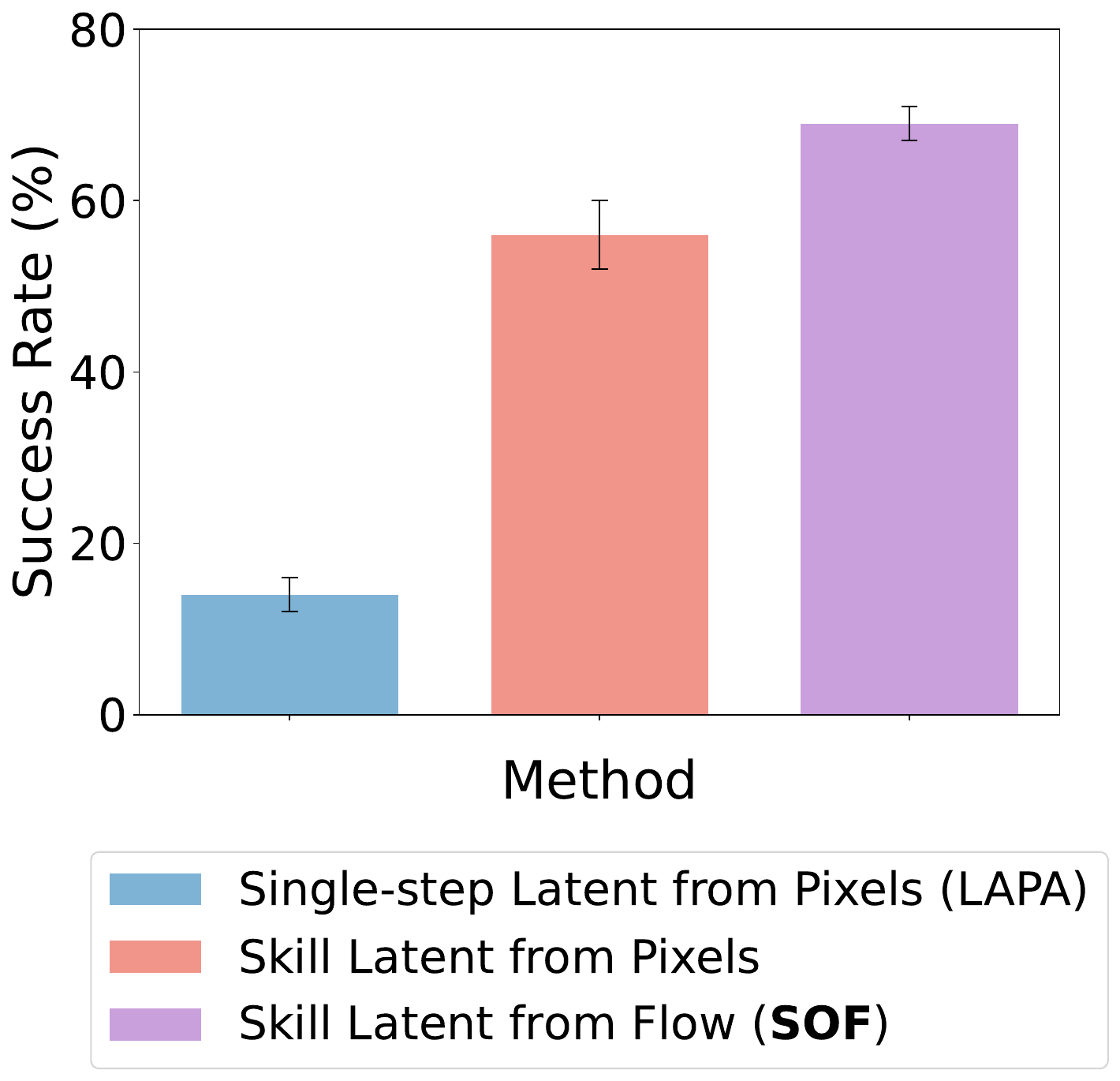}
  \caption{\textbf{Comparison of skill learning from different visual representations.} Learning from optical flow outperforms learning directly from pixel space. LAPA is a variant that learns single-step latent actions instead of temporally extended skills.}
  \label{fig:latent_skill}
\end{minipage}
\hfill
\begin{minipage}[c]{0.45\linewidth}
  \centering
  \includegraphics[width=\linewidth]{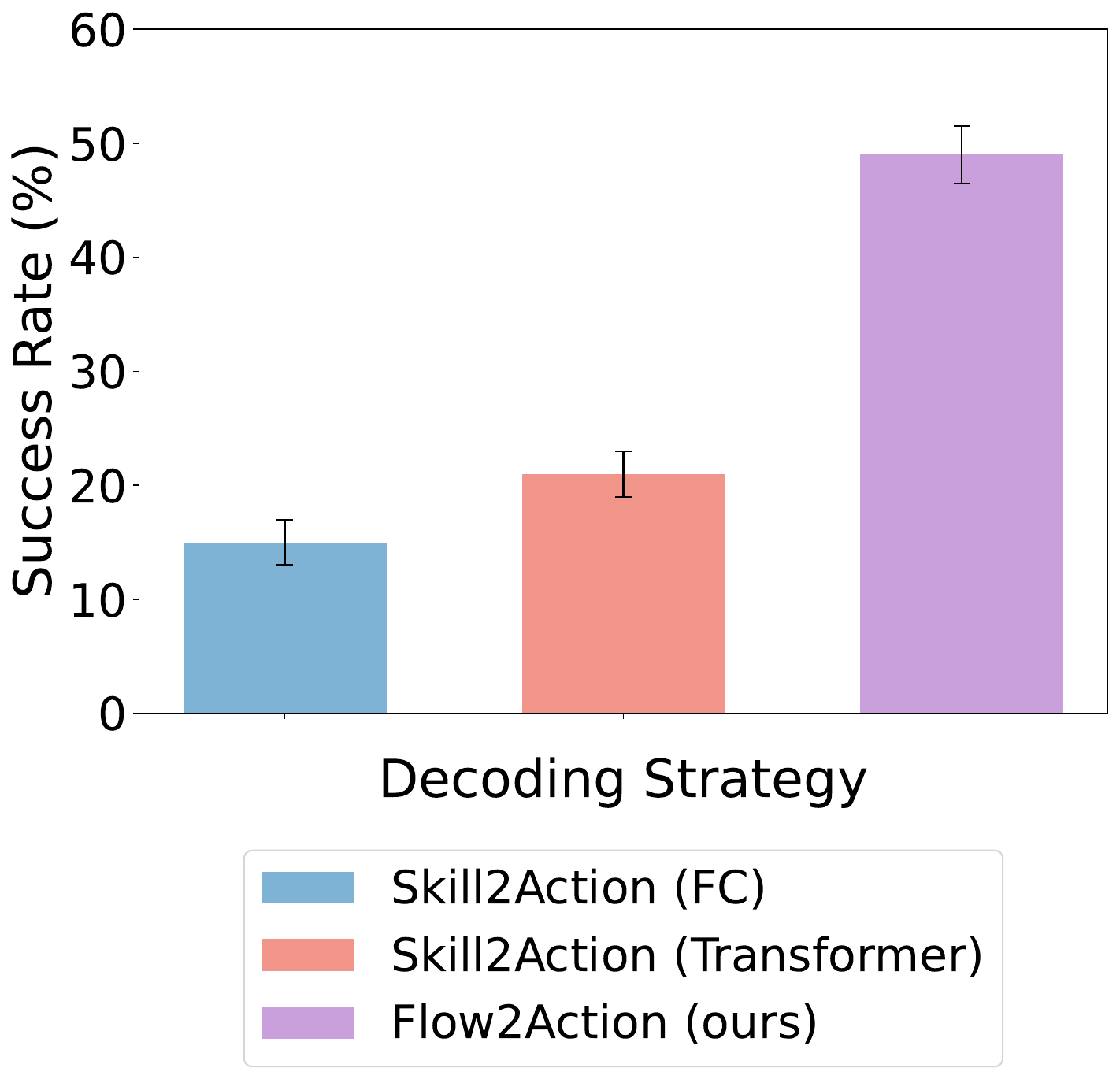}
  \caption{\textbf{Comparison of Skill-to-Action Decoding Strategies.} Directly decoding skills to actions (Skill2Action) performs worse than first transferring skills to optical flow and then to action (Flow2Action).}
  \label{fig:skill2action}
\end{minipage}

\end{figure}

\noindent\textbf{Why learn latent skills with optical flow instead of pixels?}
To verify the effectiveness of learning skills from an intermediate representation, we investigate using next-frame prediction as an alternative to optical flow for skill learning. Specifically, during the skill-learning phase, we replace the optical-flow targets $\{f_1, \cdots, f_{T-1}\}$ with the sequence of future frames $\{o_2, \cdots, o_T\}$. This setup is similar to LAPA~\citep{ye2024latent}, but here the goal is to learn \emph{skills} rather than single actions. Accordingly, we substitute the Flow2Action model with an inverse dynamics model that predicts actions based on the current and next frames. As shown in Figure~\ref{fig:latent_skill}, using next-frame prediction as an action surrogate results in a 13\% lower success rate on MetaWorld compared to \Method{}. 

\begin{figure}[t]
\centering
\begin{minipage}{0.8\textwidth}
  \centering
  \includegraphics[width=0.8\linewidth, trim={40 80 40 80},clip]{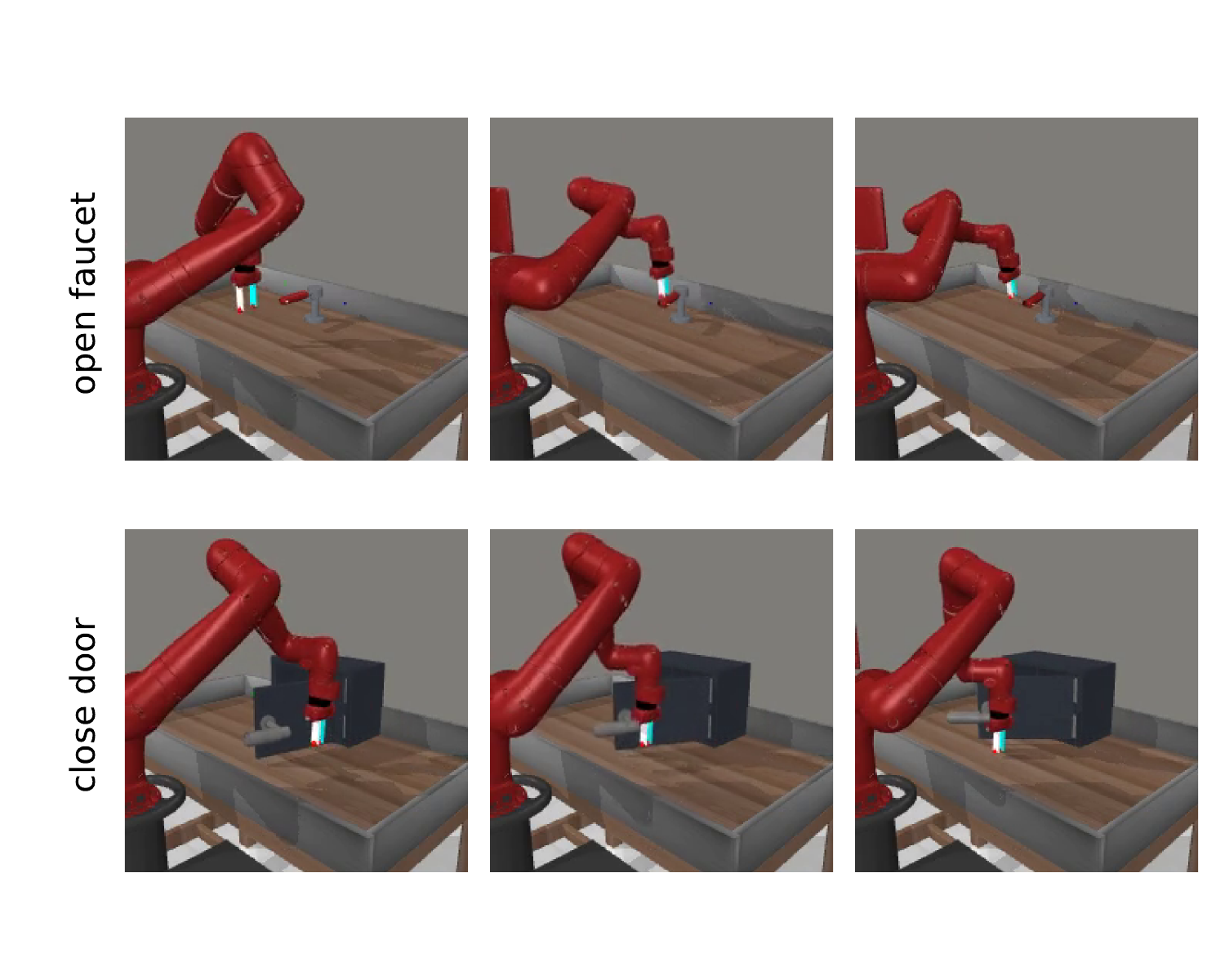}
  \captionof{figure}{\textbf{Failure cases of skill learning in pixel space.} The model can generate temporally incoherent motions when trained in pixel space.}
  \label{fig:image_based}
\end{minipage}
\label{analysis:image_based}
\end{figure}

We also analyze failure cases. As shown in Figure~\ref{fig:image_based}, the model occasionally produces contradictory motions. For example, in the open faucet task, the agent initially moves toward the faucet but then reverses direction, suggesting it may have confused the “open faucet” skill with “close faucet” due to visually similar cues. A similar issue arises in the close door task: the agent first approaches the door as if to close it but then performs a motion in the opposite direction, resembling an attempt to open the door. These cases highlight the advantages of learning in flow space, where the model focuses on motion rather than appearance.

\noindent\textbf{Why decode to flows for actions instead of decoding skills directly to actions?}
 \label{sec:skill2action}
An alternative to our design is to bypass optical flow decoding and directly map discrete skill tokens into low-level actions. We implemented two such variants: (i) a fully-connected (FC) head and (ii) a transformer decoder conditioned on the input image. Figure~\ref{fig:skill2action} summarizes the results. Both direct mapping approaches underperform significantly, with average success rates of 0.15 and 0.21, compared to 0.49 achieved by our flow-based Flow2Action module. Although direct mapping is computationally lighter, the intermediate flow representation provides a strong motion-centric prior that guides action inference, improving generalization across tasks. This suggests that optical flow serves as a valuable structured intermediate signal, and the slight overhead introduced by decoding into flow is a worthwhile trade-off for substantially higher performance.

\begin{figure}[t]
\centering

\begin{minipage}[c]{0.5\linewidth}
  \centering
  \includegraphics[width=\linewidth]{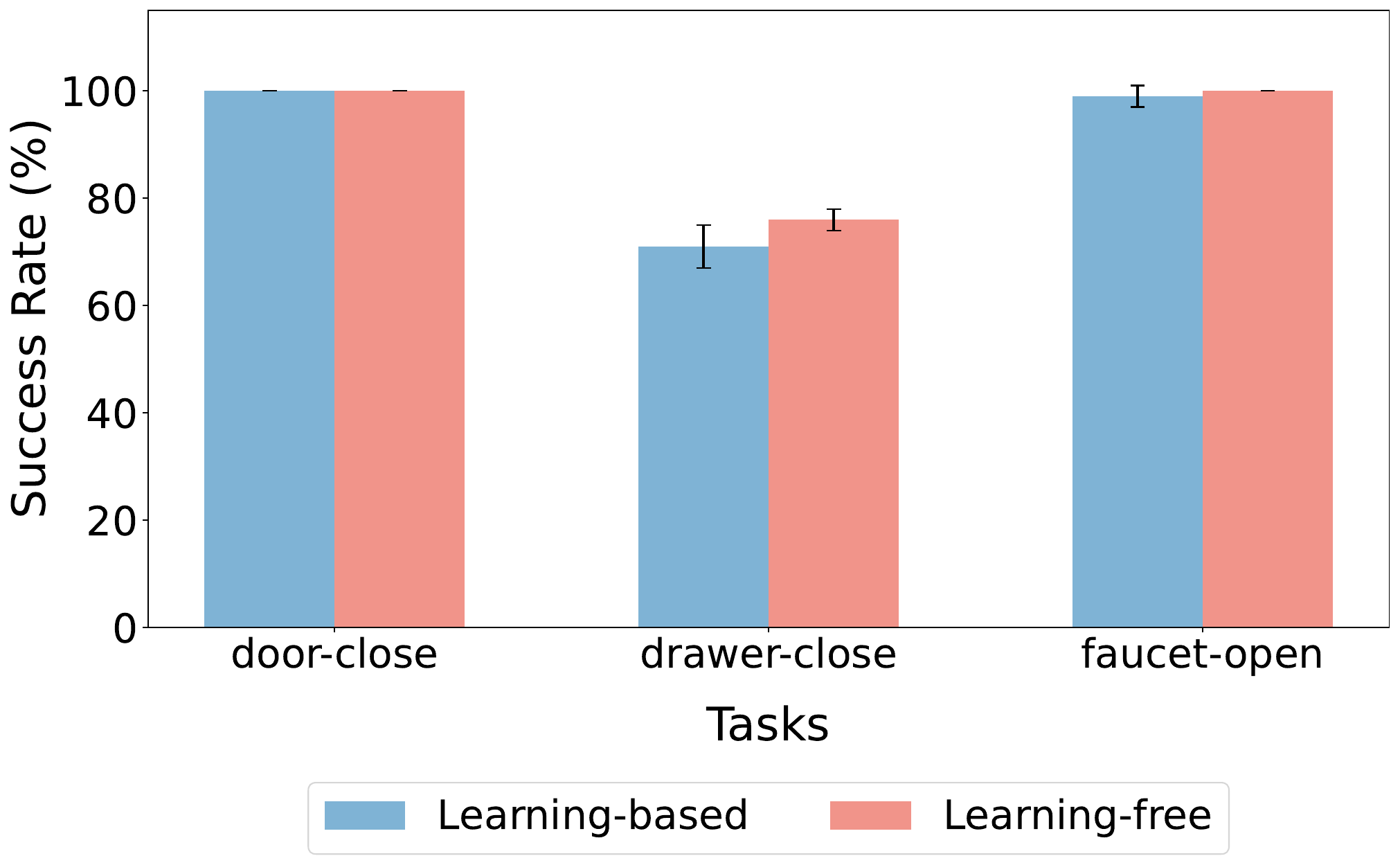}
  \caption{\textbf{Learning vs. Learning-free.} The Flow2Action module is flexible and can be implemented either by learning from action-labeled data or by using a learning-free approach without explicit action labels.}
  \label{fig:learning_free}
\end{minipage}
\hfill
\begin{minipage}[c]{0.42\linewidth}
  \centering
  \includegraphics[width=\linewidth]{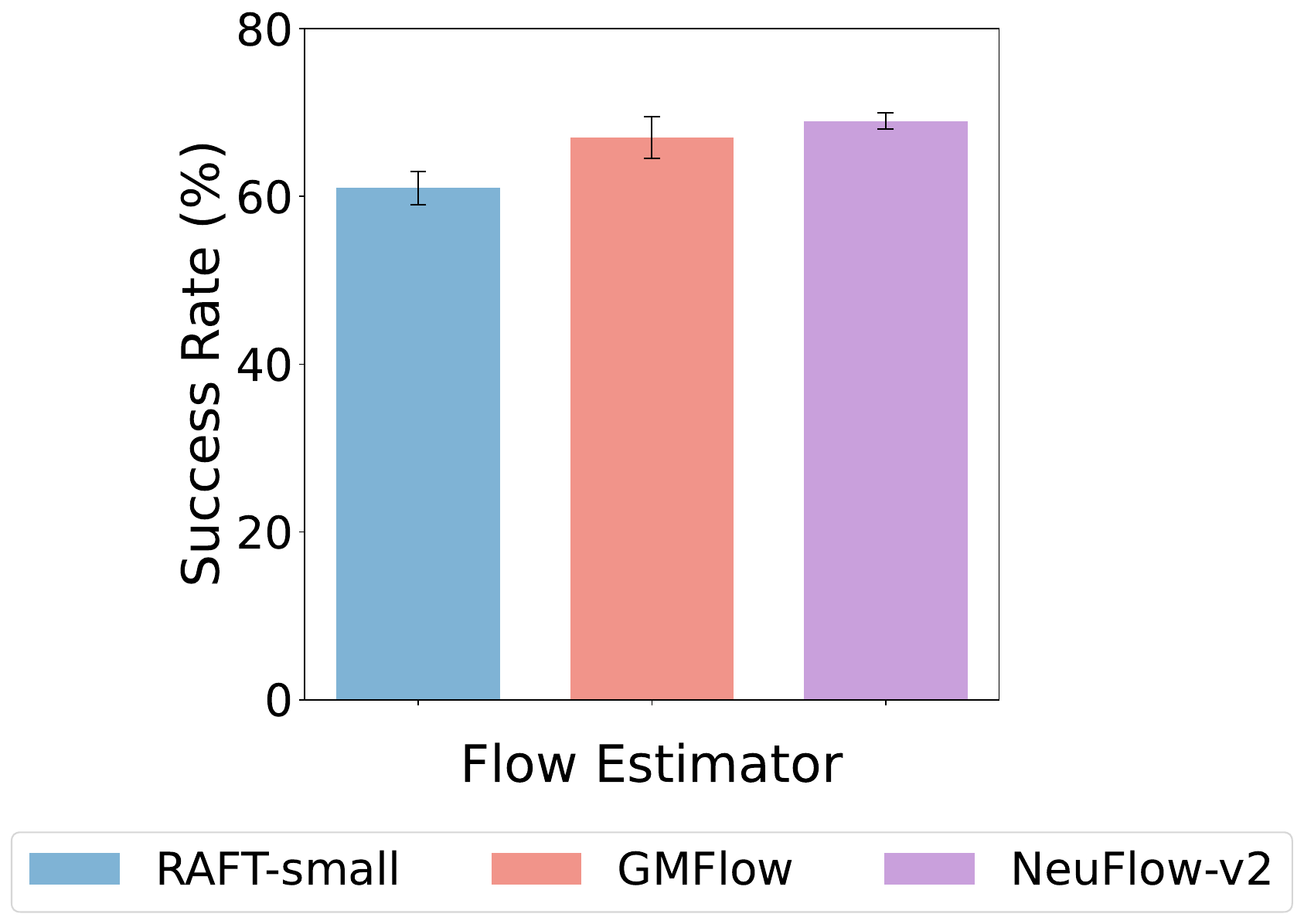}
  \caption{\textbf{Performance comparison across different optical flow estimators.} The performance is robust to different optical flow estimators, despite variations in their quality.}
  \label{fig:flow_ablation}
\end{minipage}

\end{figure}

\noindent\textbf{Learning vs. Learning-free.}
The Flow2Action module can be implemented either by training an end-to-end model that maps optical flow to actions or by using learning-free methods such as AVDC~\citep{ko2023learning}. In Figure~\ref{fig:learning_free}, we compare the performance of both approaches in MetaWorld. The results show that AVDC consistently outperforms the learning-based method. However, AVDC requires additional inputs (e.g., depth, segmentation) and prior knowledge about the environment to implement correctly. In our main experiments, we adopt the learning-based approach. Nonetheless, our Flow2Action module remains flexible, as it can be instantiated with either a learned model or a learning-free method depending on the application scenario.

\noindent\textbf{Flow estimator ablation.}
\label{sec:flow_ablation}
To evaluate the robustness of SOF to different optical flow estimators, we compared RAFT-small~\citep{eslami2024rethinkingraftefficientoptical}, GMFlow~\citep{xu2022gmflowlearningopticalflow}, and the more recent NeuFlow-v2~\cite{zhang2024neuflow}. Figure~\ref{fig:flow_ablation} reports the average success rates on MetaWorld tasks. While performance degrades with weaker flow models, our framework remains competitive. In particular, GMFlow achieves results close to NeuFlow-v2 despite exhibiting slight background noise, indicating that SOF is resilient to moderate estimation errors. In contrast, RAFT-small shows a noticeable drop in success, likely due to its limited capacity, yet it still surpasses baseline methods. These results demonstrate that SOF is not overly reliant on a specific flow estimator and can adapt across multiple choices. 


\section{Conclusion}
\label{sec:conclusion}

We introduced \Method{}, a framework for learning composable and transferable robotic skills directly from action-free videos by leveraging optical flow as a surrogate for action. \Method{} extracts structured motion primitives from raw videos, enabling policy learning in the learned skill space and translating these plans into executable actions via both learning-based and learning-free modules. Our experiments across multi-task, long-horizon, and cross-embodiment settings demonstrate that \Method{} improves performance over prior learning-from-video methods while requiring only minimal action supervision. The results highlight the potential of mid-level motion representations for scalable robot learning and open new directions for skill discovery from unstructured visual data. 

\myparagraph{Limitation and future work}
Future work may extend \Method{} to broader data sources, such as human and egocentric videos, enabling more scalable and diverse skill learning beyond robot videos. In addition, our reliance on flow introduces certain limitations, such as occlusions between the robot arm and objects, sensitivity to visual instability, and dependence on fixed camera positions. To address these challenges, future work may explore alternative representations as action surrogates, such as extending to 3D using scene flow. We also aim to deploy our method in real-world settings to assess its practical applicability.

\section{Ethics statement}
Our work focuses on improving the scalability and generalization of robot learning from unlabeled videos, which can benefit applications such as assistive robotics, home automation, and industrial manipulation. Since our method builds on publicly available datasets and models, and does not involve human subjects or sensitive data, we do not foresee any obvious negative societal impacts. Nonetheless, we encourage responsible use and emphasize that our framework should be applied in alignment with safety and ethical guidelines.

 \section{Reproducibility statement}

We have included the implementation details, training setup, training time, and hardware specifications in Appendix~\ref{appendix:exp_details} and~\ref{appendix:compute} to ensure reproducibility.

\bibliography{reference}
\bibliographystyle{iclr2026_conference}

\appendix
\appendix

\begingroup
\hypersetup{colorlinks=false, linkcolor=black}
\hypersetup{pdfborder={0 0 0}}
\part{} 
\parttoc 
\endgroup

\section{Experiment Details}
\label{appendix:exp_details}

\subsection{Training hyperparameters}

The hyperparameters of all training stages of \Method{} are listed in Table~\ref{table:hyperparameter}.

\begin{table}[ht]
\centering
\small
\caption[Training hyperparameters]{\textbf{Training hyperparameters.}}
\vspace{5pt}
\scalebox{0.75}{
\begin{tabular}{@{}ccc@{}}\toprule
\textbf{Stage} & \textbf{Hyperparameter} & value \\
\cmidrule{1-3}
\multirow{3}{*}{Stage 1} 
& Encoder Dim. & 256          \\
& Eecoder Dim. & 256          \\
& Skill block size &  32      \\
& Downsample factor &  4      \\
& Attn. Dropout &   0.1    \\
& Encoder heads &  4   \\
& Encoder layers &  2   \\
& Decoder heads &   4   \\
& Decoder layers &   4   \\
& VQ type &  fsq   \\
& Codebook Size & 1024 \\
& Learning Rate & 0.0001 \\
& Batch Size  & 256 \\
\cmidrule{1-3}
\multirow{3}{*}{Stage 2} 
& N layers & 6 \\
& N heads & 6 \\
& Embedding Dim. & 384 \\
& Attn. Dropout & 0.1 \\
& Embedding Dropout & 0.1\\ 
& Beam size & 5 \\
& Temperature & 1.0 \\
& Learning Rate & 0.0001 \\
& Batch Size  & 128 \\
\cmidrule{1-3}
\multirow{3}{*}{Stage 3} 
& Base model & resnet18 \\
& Learning Rate & 0.0001 \\
& Batch Size  & 128 \\
\bottomrule
\end{tabular}
}
\label{table:hyperparameter}
\end{table}

\subsection{Implementation details of baselines}



\noindent\textbf{AVDC.} We follow the codebase\footnote{\url{https://github.com/flow-diffusion/AVDC}} to train the video model. To fit within the memory constraints of a single 24GB GPU, we reduce the batch size to 2. For converting generated videos into executable actions, we adopt the approach provided in the official codebase\footnote{\url{https://github.com/flow-diffusion/AVDC_experiments/tree/main}}, which utilizes optical flow, depth, and segmentation masks. We evaluate the results only on MetaWorld, as we were unable to get the action transformation pipeline to work in the LIBERO environment due to the need for additional environment-specific design. Recent work~\citep{luo2025grounding} also reports near-zero success rates on LIBERO.

\noindent\textbf{LAPA.} We follow the official codebase\footnote{\url{https://github.com/LatentActionPretraining/LAPA}} to train the model on MetaWorld and LIBERO. During the latent pretraining stage, we reduce the number of training steps to 1,000, given the small-scale datasets with only 50 demonstrations per task. In the action fine-tuning stage, we fine-tune the model for 500 steps. All experiments are conducted using 8 V100 GPUs.

\subsection{Cross-embodiment tasks}
\label{appendix:cross-embodiment-tasks}

The task sets used in setting \textbf{(b)} of cross-embodiment transfer are listed in Table \ref{table:cross-embodiment-tasks}.

\begin{table}[ht]
\centering
\caption[Cross-embodiment transfer task assignment]{\textbf{Cross-embodiment transfer task assignment.}}
\vspace{5pt}
\scalebox{0.9}{
\begin{tabular}{c c c c c c}\toprule
\textbf{Task set} & & & & & \\
\midrule
\textbf{Set A} & door-open & door-close & basketball & hammer & button-press-topdown \\
\midrule
\textbf{Set B} & faucet-close & faucet-open & handle-press & button-press & assembly \\
\bottomrule
\end{tabular}
}
\label{table:cross-embodiment-tasks}
\end{table}


\section{Computational Resources}
\label{appendix:compute}

We use the workstations listed in Table \ref{table:computational-resouces}. 
Our method requires approximately 20 hours for the first stage, 3 hours for the second stage, and 3 hours for the third stage, totaling around 26 GPU hours on a single workstation. 
For reference, the training cost of comparable baselines is on a similar scale: behavior cloning (BC) takes about 1 hour for 30 demonstrations, Diffusion Policy requires roughly 4 hours for 30 demonstrations, AVDC uses approximately 24 GPU hours in total (12 hours for video diffusion training on 2 RTX 4090 GPUs plus a training-free Flow2Action module), and LAPA requires about 30 GPU hours (3 hours for latent action quantization on 2 V100s, 2 hours for latent pretraining on 8 V100s, and 1 hour for finetuning on 8 V100s). 
Overall, the computational cost of our method is comparable to or slightly lower than recent action-free video pretraining approaches, while delivering consistent improvements across tasks and benchmarks.

\noindent\textbf{Latency Analysis.}
We measure end to end control frequency on a single RTX 4090. The SOF pipeline runs at about 15 Hz from vision input to predicted skill tokens to decoded flow to actions. Under the same setup, Diffusion Policy runs at about 10 Hz.

\begin{table*}[!ht]
\centering
\caption[Computational Resources]{\textbf{Computational resources.}}
\scalebox{0.9}{\begin{tabular}{@{}cccc@{}}\toprule
\textbf{Workstation} 
& CPU
& GPU
& RAM
\\
\cmidrule{1-4}
Workstation 1
& Intel Xeon w7-2475X
& NVIDIA GeForce RTX 4090 x 2
& 125 GiB
\\
Workstation 2
& Intel Xeon w5-2455X
& NVIDIA RTX A6000 x 2
& 125 GiB
\\
Workstation 3
& Intel Xeon W-2255
& NVIDIA GeForce RTX 4070 Ti x 2
& 125 GiB
\\
\bottomrule
\end{tabular}
}
\label{table:computational-resouces}
\end{table*}

\end{document}